\documentclass[10pt,twocolumn,letterpaper]{article}

\usepackage{cvpr}
\usepackage{times}
\usepackage{epsfig}
\usepackage{graphicx}
\usepackage{amsmath, mathtools}
\usepackage{amssymb}

\usepackage{multirow}
\usepackage{makecell}
\usepackage{color}
\usepackage[dvipsnames]{xcolor}
\usepackage[font=normal]{subfig}
\usepackage[export]{adjustbox}
\usepackage{setspace}
\usepackage{booktabs}
\usepackage{array}

\usepackage[pagebackref=true,breaklinks=true,letterpaper=true,colorlinks,bookmarks=false]{hyperref}

\cvprfinalcopy 

\def\cvprPaperID{****} 

\ifcvprfinal\fi

\begin{document}

\title{NH-HAZE: An Image Dehazing Benchmark with Non-Homogeneous Hazy and Haze-Free Images}

\author{Codruta O. Ancuti$^{*}$, Cosmin Ancuti$^{*\dag}$ and Radu Timofte$^{\ddag}$\\
$^{*}$Universitatea Politehnica Timisoara, Romania\\
$^{\dag}$Institute of Informatics and Applications, University of Girona, Spain\\
$^{\ddag}$ETH Zurich, Switzerland
}

\maketitle

\begin{abstract}
Image dehazing is an ill-posed problem that has been extensively studied in the recent years. The objective performance evaluation of the dehazing methods is one of the major obstacles due to the lacking of a reference dataset. While the synthetic datasets have shown important limitations, the few realistic datasets introduced recently assume homogeneous haze over the entire scene.
Since in many real cases haze is not uniformly distributed  we introduce \textbf{NH-HAZE}, a non-homogeneous realistic dataset with pairs of real hazy and corresponding haze-free images. This is the first non-homogeneous image dehazing dataset and contains 55 outdoor scenes. The non-homogeneous haze has been introduced in the scene using a professional haze generator that imitates the real conditions of hazy scenes. Additionally, this work presents an objective assessment of several state-of-the-art single image dehazing methods that were evaluated using \textbf{NH-HAZE} dataset. 
\end{abstract}


\section{Introduction}

Haze is an atmospheric phenomenon that highly influences the quality of images captured under such conditions. In consequence, haze may reduce the performance of various outdoor applications. Haze is characterized by a high density of floating particle in the air which reduces significantly the image quality in terms of contrast and color shifting.

\begin{figure*}[ht!]
  \centering
  \includegraphics[width=1\linewidth]{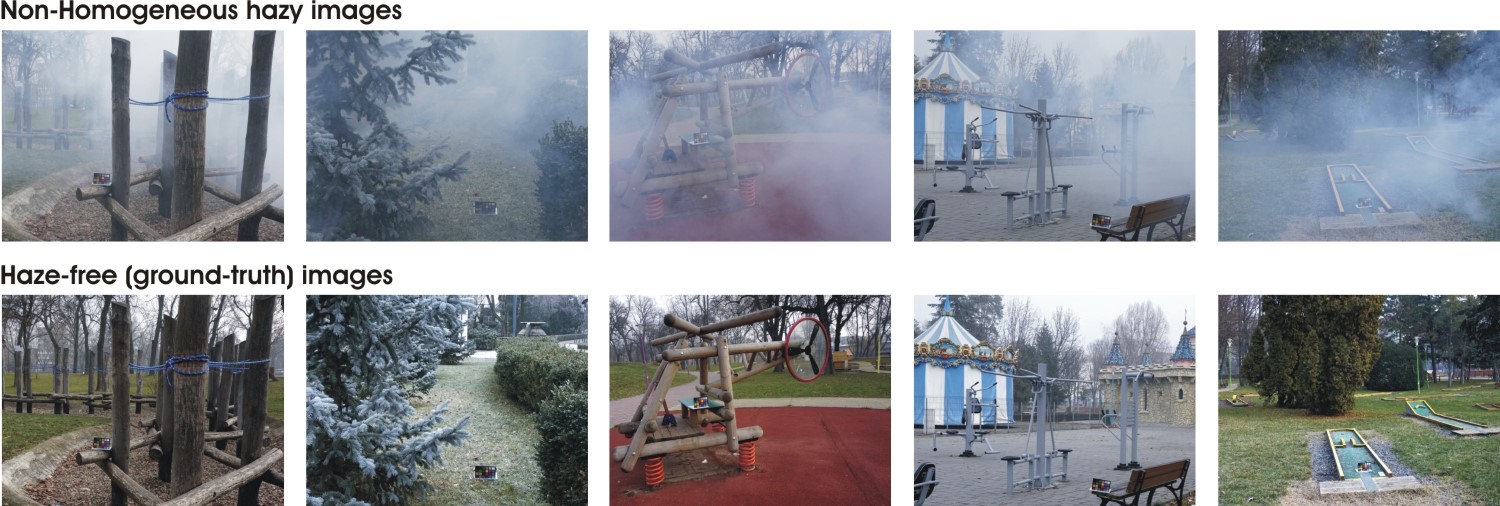}
  \caption{\label{fig:res_comp1}%
    \textit{\textbf{NH-HAZE dataset.}} Five sets of images of the \textbf{NH-HAZE} dataset.
  }  
\end{figure*}

Image dehazing aims at restoration of image contents affected by the haze. This is an ill-posed problem that has been solved initially using additional information~\cite{Cozman_Krotkov_97,Schechner_2003}. Most of the early single image dehazing methods solve the optical model of Koschmieder~\cite{Koschmieder_1924} by searching for different priors that capture statistical properties of the hazy and non-hazy images~\cite{Fattal_Dehazing,Tan_Dehazing,Dehaze_He_CVPR_2009,Tarel_ICCV_2009,Kratz_and_Nishino_2009,Dehaze_Ancuti_ACCV,Ancuti_TIP_2013,Fattal_Dehazing_TOG2014,Emberton_2015,Tang_2014}. 

One of the major obstacles in image dehazing is the validation of the proposed methods. Despite of their large number and variety, the quality of the image dehazing solutions is in many times debatable. Therefore, identifying their limitations and the new research directions is quite difficult. The image dehazing field is still lacking standardized benchmarks allowing objective and quantitative performance evaluation of the dehazing techniques. Basically, a major issue to objectively assess image dehazing performance is the absence of reference haze-free images. Collecting pairs of hazy and corresponding haze-free (ground-truth) images is a burdensome task since both images (haze and haze-free) are required to be captured under identical conditions.
  
Recently, important steps forward have been made by image dehazing challenges organized in conjunction with the IEEE CVPR NTIRE workshops~\cite{Ancuti_NTIRE_2018,Ancuti_NTIRE_2019}.
The NTIRE challenges employed new realistic image dehazing datasets (with haze and haze-free images): I-HAZE~\cite{Ancuti_IHAZE_2018}, O-HAZE~\cite{Ancuti_OHAZE_2018} and DENSE-HAZE~\cite{Ancuti_DENSE_HAZE_2019}. 

One limitation of these datasets is the common assumption that the haze is homogeneously distributed over the entire scene. In fact, haze distribution has a non-homogeneous character in many real scenes. Therefore, the existence of a dataset with non-homogeneous haze is very important for the image dehazing field.

This work introduces  \textbf{NH-HAZE}\footnote{\scriptsize{\url{https://data.vision.ee.ethz.ch/cvl/ntire20/nh-haze/}}} which represents the first realistic image dehazing dataset with non-homogeneous hazy and haze-free (ground-truth) paired images. The non-homogeneous haze has been generated using a professional haze generator that imitates the real conditions of haze scenes. \textbf{NH-HAZE} contains 55 pairs of images recorded outdoor. Our new dataset has been employed by the IEEE CVPR 2020 NTIRE workshop associated challenge on image dehazing~\cite{Ancuti_NTIRE_2020}, a challenge which attracted hundreds of registered participants. 

Additionally, this work presents a comprehensive evaluation of several state-of-the-art single image dehazing methods, that were objectively evaluated on our new dataset. Since \textbf{NH-HAZE} dataset contains ground-truth (haze-free) images, the analyzed single image dehazing techniques have been assessed quantitatively using two traditional metrics: PSNR and SSIM~\cite{Wang_2004}.

\section{Related Work}
\label{sec:related_work}

\subsection{Image dehazing methods}
\label{ssc:image_dehazing_methods}

Image dehazing is an ill-posed problem that has been solved initially based on additional information such as  atmospheric cues~\cite{Cozman_Krotkov_97,Narasimhan_2002}, multiple images captured with polarization filters~\cite{PAMI_2003_Narasimhan_Nayar,Schechner_2003}, or known depth information~\cite{Kopf_DeepPhoto_SggAsia2008,Tarel_ICCV_2009}. 

More recently, single image dehazing techniques employ the optical model of Koschmieder~\cite{Koschmieder_1924} searching for different priors that capture statistical properties of the hazy and non-hazy images~\cite{Fattal_Dehazing,Tan_Dehazing,Dehaze_He_CVPR_2009,Tarel_ICCV_2009,Kratz_and_Nishino_2009,Dehaze_Ancuti_ACCV,Ancuti_TIP_2013,Fattal_Dehazing_TOG2014,Emberton_2015,Tang_2014}. Tan~\cite{Tan_Dehazing} optimizes the local contrast based on the observation that the airlight highly influences the edge information of hazy images. Dark channel prior DCP~\cite{Dehaze_He_CVPR_2009} is based on the assumption  that in non-hazy regions without sky, the intensity value of at least one color channel within a local window is close to zero. Color lines~\cite{Fattal_Dehazing_TOG2014} and haze-line~\cite{Berman_2016} priors were built on the observation that the color distribution impacts the color channels distribution. Color channel compensation~\cite{Ancuti_3C_TIP_2020} exploits the observation that color images taken under extreme illumination present low intensity of at least one color channel. 

Another direction of research in image dehazing includes those methods that restore the visibility of hazy image without assuming the optical model~\cite{Tan_Dehazing,Ancuti_TIP_2013,Choi_2015}. For instance local contrast maximization-based methods~\cite{Tan_Dehazing,Tarel_ICCV_2009} and fusion-based techniques~\cite{Ancuti_TIP_2013,Choi_2015,Ancuti_NightTime,Ancuti_NT_TIP_2020} shown effectiveness for single image dehazing, without an explicit transmission estimation.

The advent of and the advances in the field of deep-learning led also to competitive learning-based solutions for image dehazing. \textit{DehazeNet}~\cite{Dehazenet_2016} takes a hazy image as input and outputs its medium transmission map that is subsequently used to recover a haze-free image via atmospheric scattering model. For its training, \textit{DehazeNet} resorts to data that is synthesized based on the physical haze formation model. Ren~\etal~\cite{Ren_2016} proposed a coarse-to-fine network consisting of a cascade of convolutional neural network (CNN) layers, also trained with synthesized hazy images. For a diverse selection of deep learned solutions we refer the reader to the recent NTIRE dehazing challenges reports~\cite{Ancuti_NTIRE_2018,Ancuti_NTIRE_2019, Ancuti_NTIRE_2020}.

\subsection{Dehazing assessment}
\label{ssc:dehazing_assessment}

Although the progress made in image dehazing is remarkable, an important problem remains: the evaluation of the proposed methods. Objective assessment of the dehazing performance of a given algorithm was limited due to the absence of reference haze-free images (ground-truth). Collecting pairs of hazy and corresponding haze-free images is a burdensome task since both images (hazy and haze-free) are required to be captured under identical conditions.

Due to this limitation, earlier dehazing quality metrics were restricted to non-reference image quality metrics (NR-IQA)~\cite{Mittal_2012,Mittal_2013,Saad_2012}. Hautiere~\etal~\cite{Hautiere_2008} simply relied on the gradient of the visible edges. Chen~\etal~\cite{Chen_2014} employed a subjective assessment of enhanced and original images captured in bad visibility conditions. Choi~\etal~\cite{Choi_2015} introduced Fog Aware Density Evaluator (FADE), a blind measure, which aims to predict the visibility of a hazy scene from a single image without using a haze-free (reference) image. Unfortunately, in the absence of a ground-truth image, these blind image dehazing assessment techniques are not very accurate and therefore have not been generally accepted by the dehazing community.

A more successful strategy builds upon synthesized hazy images. The synthetic hazy images have been generated assuming the simplified optical model. As a result, considering an image with known depth map (related with the transmission map of the optical model) the haze effect is synthesized straightforwardly. Tarel~\etal~\cite{Tarel_2012} introduced FRIDA, one of the first synthetic image dehazing datasets. FRIDA contains 66 pairs of images with the hazy scenes generated using computer graphics. D-Hazy~\cite{D_Hazy_2016} dataset uses the images and the depth maps of the \textit{Middleburry}\footnote{\scriptsize{\url{http://vision.middlebury.edu/stereo/data/scenes2014/}}} and the \textit{NYU-Depth V2}\footnote{\scriptsize{\url{http://cs.nyu.edu/~silberman/datasets/nyu_depth_v2.html}}} datasets. The haze is synthesized based on Koschmieder's optical model~\cite{Koschmieder_1924} assuming  a pure white value of the airlight constant.

RGB-NIR~\cite{RGB_NIR_2017} is a relatively small dataset that contains only 4 sets of hazy, haze-free and NIR ground-truth indoor images. 

The O-HAZE~\cite{Ancuti_OHAZE_2018} is the first introduced realistic dataset that contains hazy and haze-free (ground-truth) images. It consists of $45$ various outdoor scenes captured using a professional haze generator under controlled illumination. I-HAZE~\cite{Ancuti_IHAZE_2018} dataset is similar to O-HAZE but recorded in indoor environments. I-HAZE and O-HAZE were employed by the first image dehazing challenge~\cite{Ancuti_NTIRE_2018} organized in conjunction with the 2018 IEEE CVPR NTIRE workshop\footnote{\scriptsize{\url{www.vision.ee.ethz.ch/ntire18/}}}.
While O-HAZE and I-HAZE consists of relatively light and homogeneous haze, DENSE-HAZE~\cite{Ancuti_DENSE_HAZE_2019} is a realistic dataset that contains dense (homogeneous) hazy and haze-free (ground-truth) images. DENSE-HAZE was employed by the image dehazing challenge~\cite{Ancuti_NTIRE_2019} at the 2019 IEEE CVPR NTIRE workshop.

Complementary to prior work, in this paper we introduce \textbf{NH-HAZE}, the first realistic image dehazing dataset with non-homogeneous hazy and haze-free (ground-truth) images. 

\section{Recording the NH-HAZE dataset}
\label{sec:recording}

\textbf{NH-Haze} dataset contains 55 various outdoor scenes captured with and without haze. \textbf{NH-Haze} is the first dehazing dataset that contains non-homogeneous haze scenes. Our dataset allows to investigate the contribution of the haze over the scene visibility by analyzing the scene objects radiance starting from the camera proximity to a maximum distance of 20-30m.

    The recording outdoor conditions had to be similar to the ones encountered in hazy days and therefore the recording period has been spread over more than two months during the autumn season. Basically, all outdoor scenes have been recorded during cloudy days, in the morning or in the sunset. We also had to deal with the wind speed. In order to limit fast spreading of the haze in the scene, the wind during recording had to be below 2-3 km/h. The absence of wind criterion was the hardest to satisfy and explains the long recording duration.

   The hardware used to record the scenes consisted from a tripod and a Sony A5000 camera remotely controlled (Sony RM-VPR1). We recorded JPG and ARW (RAW) 5456$\times$3632 images, with 24 bit depth. Each scene acquisition started with manual adjustment of the camera settings. The shutter-speed (exposure-time), the aperture (F-stop),  the ISO and white-balance parameters have been set at the same level when capturing the haze-free and hazy scene. 
   
   To set the camera parameters (aperture-exposure-ISO), we used an external exponometer (Sekonic) while for setting the white-balance, we used the middle gray card (18\% gray) of the color checker. For this step we changed the  camera white-balance mode in manual mode and placed the reference grey-card in the front of it. 

   To introduce haze in the outdoor scenes we employed two professional haze machines (LSM1500 PRO 1500 W), which generate vapor particles with diameter size (typically 1 - 10 microns) similar to the atmospheric haze particles. The haze machines use cast or platen type aluminum heat exchangers to induce liquid evaporation. We chose special (haze) liquid with higher density in order to simulate the effect occurring with water haze over larger distances than the investigated 20-30 meters.
   
   The generation of haze took approximately 2-3 minutes. After starting to generate haze, we used a fan to spread the haze in the scene in order to reach a nonuniform distribution of the haze in a rage of 20-30 meters in front of the camera.

    Moreover, in each outdoor recorded scene a Macbeth color checker was placed to allow for post-processing. We used a classical Macbeth color checker of size 11 by 8.25 inches and a 4$\times$6 grid of painted square samples. 

\begin{figure*}[th!]
  \centering
  \includegraphics[width=1\linewidth]{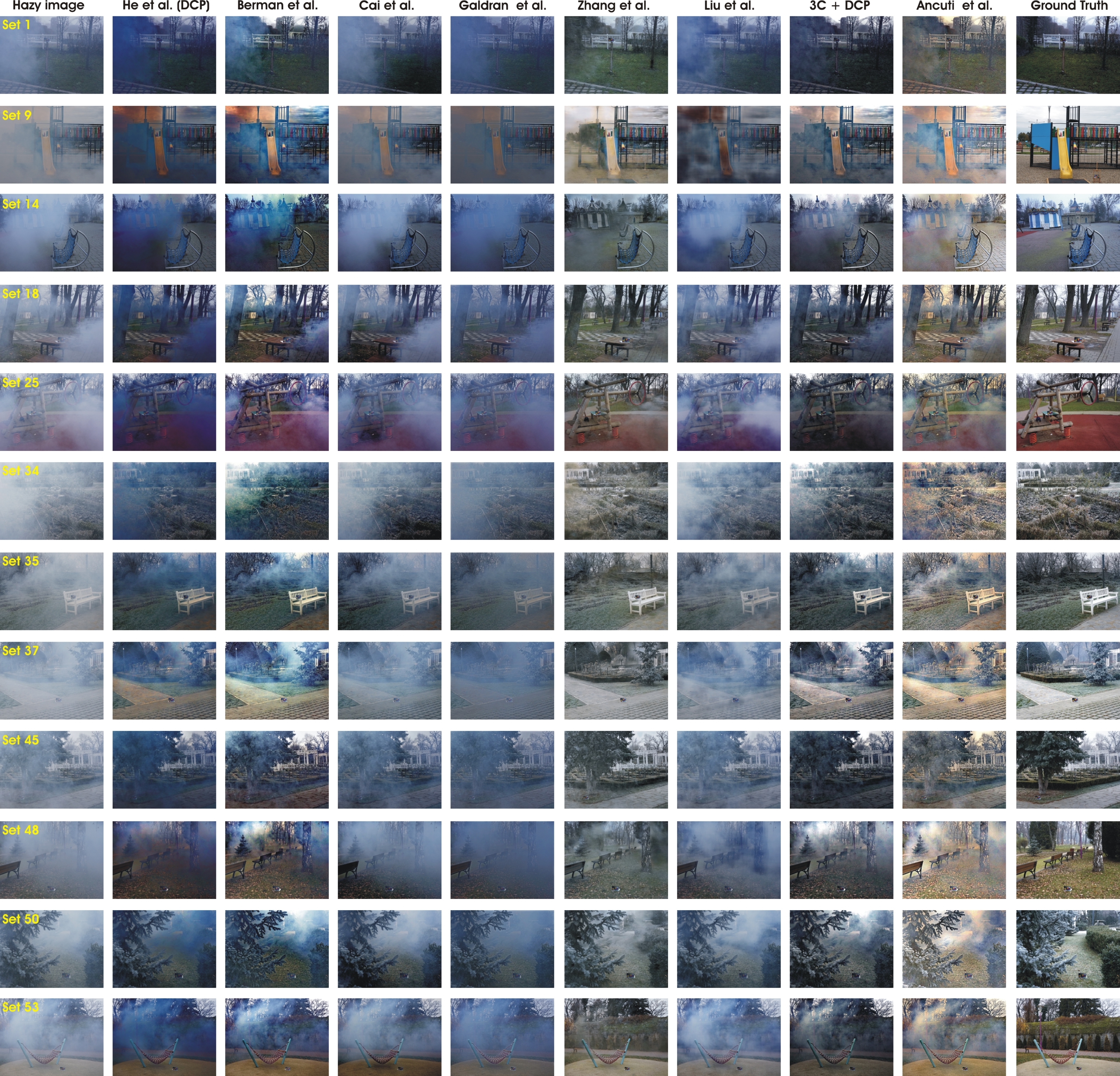}
  \caption{\label{fig:res_comp1}%
    \textit{\textbf{Comparative results.}} The first row shows the hazy images and the last row shows the ground-truth. The other rows from left to right show the results of He~\etal~\cite{Dehaze_He_CVPR_2009}, 
    Cai~\etal~\cite{Dehazenet_2016}, 
    Berman~\etal~\cite{Berman_2016}, 
    Galdran~\etal~\cite{Galdran_2018}, 
    Zhang~\etal~\cite{Zhang_dehazing_2018}, 
    Liu~\etal~\cite{GridDehazeNet_2019}, 3C~\cite{Ancuti_3C_TIP_2020} and 
    Ancuti~\etal~\cite{Ancuti_NT_TIP_2020}.
  }  
\end{figure*}

\section{Evaluated Dehazing Techniques}
\label{sec:evaluated_dehazing_techniques}

As previously mentioned, we performed a validation of several competitive image dehazing techniques based on our \textbf{NH-HAZE} dataset. We briefly discuss these image dehazing techniques in the following paragraphs. 
\newline

\textbf{He~\etal}~\cite{Dehaze_He_CVPR_2009} is one of the first single image dehazing proposed techniques. They introduced Dark Channel Prior (DCP), an extension of the dark object~\cite{Chavez_1988}. DCP has been used by many recent image dehazing techniques to estimate the transmission map of the optical model. This statistic is based on the observation that most of an outdoor image regions have pixels that present low intensity in at least one of the color channels. The exception of this rule is represented by the regions of sky and with haze. DCP helps to roughly estimate (per patch) the transmission map (directly related with the depth map of the scene). In the original work the transmission was refined by applying an expensive alpha matting strategy. In this evaluation the refinement of DCP approach was performed based on the guiding filter.   
\newline

\begin{table*}[ht!]
\centering
\resizebox{\linewidth}{!}
{
\begin{tabular}{|l|l|l|l|l|l|l|l|l|l|l|l|l|l|l|l|l|}
\hline
\multirow{2}{*}{} & \multicolumn{2}{l|}{\textbf{He~\etal (DCP)}} & \multicolumn{2}{l|}{\textbf{Berman~\etal}} & \multicolumn{2}{l|}{\textbf{Cai~\etal}} & \multicolumn{2}{l|}{\textbf{Galdran~\etal}} & \multicolumn{2}{l|}{\textbf{Zhang~\etal}} & \multicolumn{2}{l|}{\textbf{Liu~\etal}} & \multicolumn{2}{l|}{\textbf{3C+DCP}} & \multicolumn{2}{l|}{\textbf{Ancuti~\etal}} \\ \cline{2-17} 
                  & \textbf{PSNR}         & \textbf{SSIM}         & \textbf{PSNR}        & \textbf{SSIM}        & \textbf{PSNR}       & \textbf{SSIM}      & \textbf{PSNR}         & \textbf{SSIM}        & \textbf{PSNR}        & \textbf{SSIM}       & \textbf{PSNR}       & \textbf{SSIM}      & \textbf{PSNR}     & \textbf{SSIM}    & \textbf{PSNR}        & \textbf{SSIM}        \\ \hline
\textbf{Set 1}    & 18.093                & 0.657                 & 16.731               & 0.663                & 14.439              & 0.578              & 15.982                & 0.606                & 18.284               & 0.652               & 15.712              & 0.604              & 19.653            & 0.698            & 17.925               & 0.664                \\ \hline
\textbf{Set 9}    & 8.993                 & 0.430                 & 11.539               & 0.522                & 12.237              & 0.519              & 10.103                & 0.473                & 17.213               & 0.621               & 9.339               & 0.459              & 13.943            & 0.566            & 12.626               & 0.574                \\ \hline
\textbf{Set 14}   & 11.930                & 0.519                 & 12.438               & 0.505                & 14.627              & 0.602              & 13.308                & 0.586                & 16.264               & 0.668               & 15.332              & 0.633              & 15.826            & 0.619            & 15.797               & 0.628                \\ \hline
\textbf{Set 18}   & 12.739                & 0.513                 & 14.407               & 0.579                & 14.243              & 0.555              & 14.744                & 0.578                & 19.239               & 0.668               & 16.013              & 0.602              & 14.431            & 0.558            & 17.824               & 0.643                \\ \hline
\textbf{Set 25}   & 15.175                & 0.593                 & 17.835               & 0.691                & 16.128              & 0.637              & 17.192                & 0.649                & 20.498               & 0.712               & 15.377              & 0.642              & 15.598            & 0.601            & 18.810               & 0.697                \\ \hline
\textbf{Set 34}   & 14.627                & 0.439                 & 14.078               & 0.550                & 14.106              & 0.401              & 15.426                & 0.416                & 18.083               & 0.548               & 13.968              & 0.421              & 16.129            & 0.577            & 15.996               & 0.657                \\ \hline
\textbf{Set 35}   & 14.064                & 0.455                 & 14.319               & 0.611                & 13.307              & 0.454              & 14.054                & 0.432                & 18.883               & 0.582               & 14.111              & 0.454              & 14.938            & 0.592            & 14.756               & 0.671                \\ \hline
\textbf{Set 37}   & 12.304                & 0.580                 & 14.777               & 0.684                & 12.911              & 0.536              & 11.819                & 0.489                & 16.472               & 0.645               & 14.284              & 0.536              & 14.057            & 0.655            & 15.654               & 0.675                \\ \hline
\textbf{Set 45}   & 13.397                & 0.487                 & 13.813               & 0.552                & 13.245              & 0.499              & 13.812                & 0.503                & 19.101               & 0.619               & 13.539              & 0.505              & 13.374            & 0.521            & 15.805               & 0.587                \\ \hline
\textbf{Set 48}   & 11.425                & 0.299                 & 11.626               & 0.435                & 11.333              & 0.301              & 12.603                & 0.315                & 15.973               & 0.454               & 12.338              & 0.335              & 10.841            & 0.409            & 10.564               & 0.513                \\ \hline
\textbf{Set 50}   & 13.148                & 0.416                 & 11.443               & 0.476                & 12.528              & 0.398              & 13.693                & 0.414                & 15.887               & 0.460               & 13.726              & 0.442              & 13.210            & 0.514            & 12.130               & 0.555                \\ \hline
\textbf{Set 53}   & 13.201                & 0.471                 & 13.175               & 0.555                & 10.985              & 0.438              & 13.140                & 0.483                & 18.684               & 0.592               & 12.281              & 0.492              & 12.958            & 0.524            & 13.592               & 0.587                \\ \hline
\end{tabular}
}
\caption{\label{tabel_eval1} \textit{\textbf{Quantitative evaluation.}} We randomly picked up 12 sets from the \textbf{NH-HAZE} dataset, and computed the PSNR and SSIM between the ground-truth images and the dehazed images produced by the evaluated techniques. The hazy images, ground-truth and the results are shown in Fig.\ref{fig:res_comp1}.}
\end{table*}

\begin{table*}[]
\centering
\resizebox{\linewidth}{!}
{
\begin{tabular}{|l|l|l|l|l|l|l|l|l|}
\hline
              & \textbf{DCP} & \textbf{Berman~\etal} & \textbf{Cai~\etal} & \textbf{Galdran~\etal} & \textbf{Zhang~\etal} & \textbf{Liu~\etal} & \textbf{3C+DCP} & \textbf{Ancuti~\etal} \\ \hline
\textbf{PSNR} & 12.913       & 12.464                 & 12.379              & 13.323                  & \textbf{17.081}       & 13.086              & 13.523          & 14.296                 \\ \hline
\textbf{SSIM} & 0.472        & 0.530                  & 0.455               & 0.482                   & 0.585                 & 0.498               & 0.552           & \textbf{0.602}         \\ \hline
\end{tabular}
}
\caption{\label{table_average} \textbf{\textit{Quantitative evaluation}} on all the 55 set of images of the \textbf{NH-HAZE} dataset. This table presents the average values of the PSNR and SSIM, over the entire dataset.}
\end{table*}

\textbf{Cai~\etal}~\cite{Dehazenet_2016} introduced DehazeNet, one of the first deep learned methods for image dehazing. Dehazenet is an end-to-end learned CNN that estimates the transmission map. It is trained to map hazy to haze-free patches using an synthesized hazy dataset. \textit{Dehazenet} consists from four sequential steps: features extraction, multi-scale mapping, local extrema and finally non-linear regression. \\
\newline

\textbf{Berman~\etal}~\cite{Berman_2016} solution is based on the color consistency observation introduced by Omer~\etal~\cite{Omer_2004}. This approach considers that the color distribution in a haze-free images is well approximated by a discrete set of clusters in the RGB color space. Basically, this approach assumes that the pixels in a given cluster are non-local and are spread over the entire image plane. Therefore the pixels of a hazy region are assumed to be affected differently. For hazy images these color clusters become different lines in RGB color space, named \textit{haze-lines}. The position of a pixel within the line reflects its transmission level. Based on the haze-lines the proposed method  estimates both the transmission map and haze free image.
\newline

\textbf{Galdran~\etal}~\cite{Galdran_2018} employ the Retinex theory for image dehazing problem. Their approach applies Retinex on inverted intensities of a hazy input image proving that this strategy is effective for image dehazing.
\newline

\textbf{Zhang~\etal}~\cite{Zhang_dehazing_2018} present a CNN-based approach to dehaze images. They propose a Perceptual Pyramid Deep Network that has an encoder-decoder structure. The model is learned from paired data using a combination of mean squared error and perceptual losses. This approach is the winner of the IEEE CVPR NTIRE 2018 image dehazing challenge~\cite{Ancuti_NTIRE_2018}.
\newline

\textbf{Liu~\etal}~\cite{GridDehazeNet_2019} introduce also a CNN-based approach named \textbf{GridDehazeNet}. This network consists from three main modules. The first module pre-processes the data yielding inputs with better diversity and more pertinent features. The second module, the backbone module, allows for a more efficient information exchange
across different scales. The last module post-processes the outputs in order to reduce the level of the artifacts.\\
\newline

\textbf{3C}~\cite{Ancuti_3C_TIP_2020} introduces an original  general solution (named \textbf{3C}- \textbf{C}olor \textbf{C}hannel \textbf{C}ompensation) to improve image enhancement in terms of color appearance for images characterized by severely non-uniform color spectrum distribution. It is based on the observation that, under such adverse conditions, the information contained in at least one color channel is close to completely lost, making the traditional enhancing techniques subject to noise and color shifting.  \textbf{3C} is used as a pre-processing method that reconstructs the lost channel based on the opponent color channel. In this evaluation we employ 3C as a pre-procesing step applied to the traditional DCP.
\newline

\textbf{Ancuti~\etal}~\cite{Ancuti_NT_TIP_2020} introduce the first general image dehazing method that yields competitive results for both day and night-time hazy scenes. The method is based on a novel local airlight estimation approach that allows to effectively deal with the night-time conditions characterized in general by non-uniform distribution of the light due to the multiple localized artificial sources. Multiple patch sizes are considered to generate several images. These derived images are merged based on a  multi-scale fusion strategy guided by several weight maps. 
\newline

\section{Results and Discussion}
\label{sec:results_and_discussion}

\begin{figure*}[t!]
  \centering
  \includegraphics[width=1\linewidth]{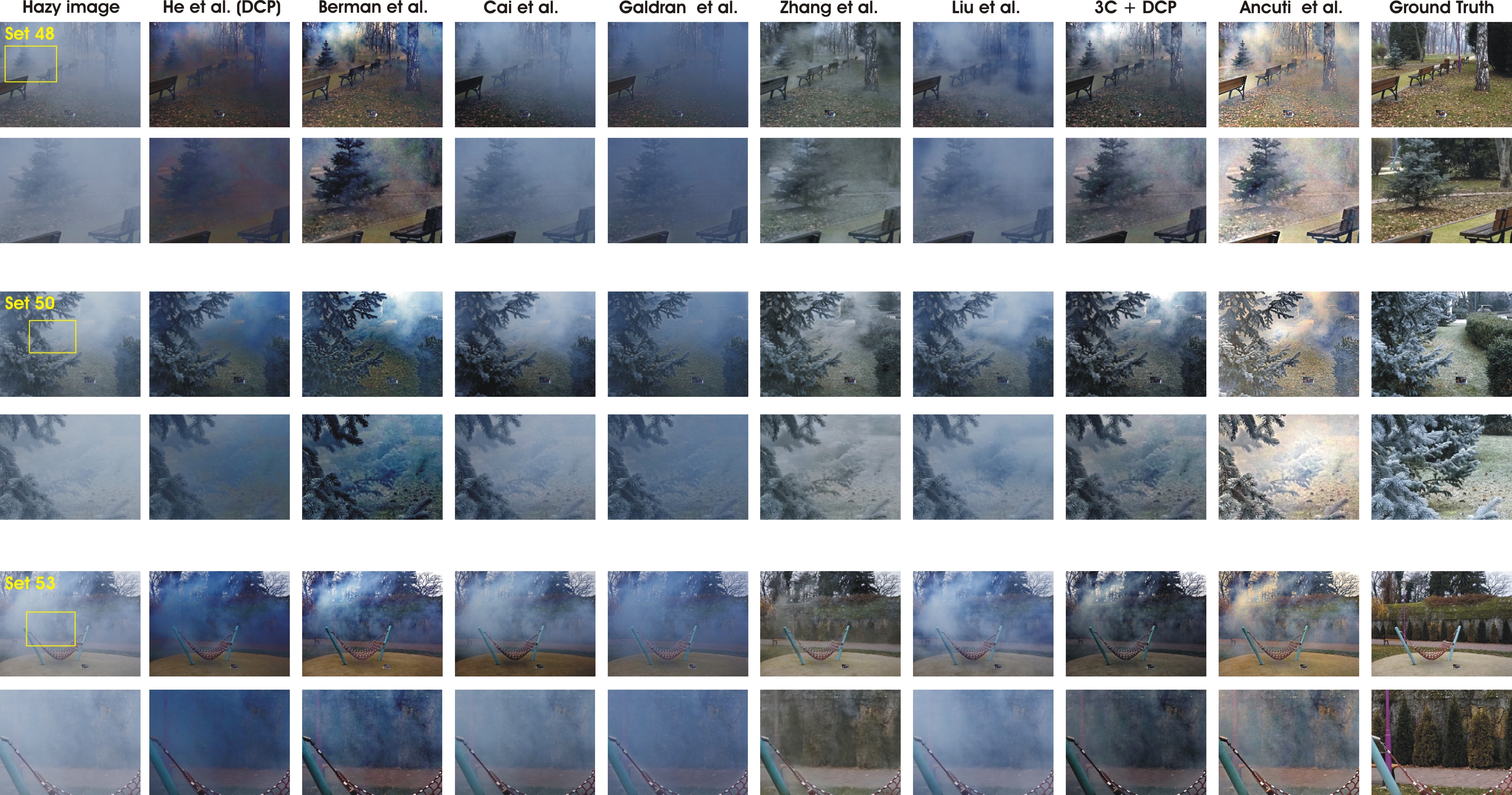}
  \caption{\label{fig:res_comp_crop1}%
    \textit{\textbf{Comparative detail insets.}} The first, third and fifth rows show  the hazy images (first column), their corresponding ground-truth (last column), and the results of several dehazing techniques
    He~\etal~\cite{Dehaze_He_CVPR_2009}, 
    Cai~\etal~\cite{Dehazenet_2016},   
    Berman~\etal~\cite{Berman_2016}, 
    Galdran~\etal~\cite{Galdran_2018}, 
    Zhang~\etal~\cite{Zhang_dehazing_2018}, 
    Liu~\etal~\cite{GridDehazeNet_2019}, 3C~\cite{Ancuti_3C_TIP_2020} and 
    Ancuti~\etal~\cite{Ancuti_NT_TIP_2020}, for three sets of the \textbf{NH-HAZE} dataset. The corresponding detail insets are shown below in the even rows.
  }  
\end{figure*}

The new \textbf{NH-HAZE} dataset has been used to perform a comprehensive evaluation of the recent competitive single image dehazing techniques presented in Section~\ref{sec:evaluated_dehazing_techniques}. We have randomly selected several images of our dataset and show them in Fig.~\ref{fig:res_comp1} (in the  first column, non-homogeneous hazy images,  and in the last column, haze free images). The other columns (from left to right) depict the results generated using the dehazing techniques of He~\etal~\cite{Dehaze_He_CVPR_2009}, 
Cai~\etal~\cite{Dehazenet_2016},   
Berman~\etal~\cite{Berman_2016}, 
Galdran~\etal~\cite{Galdran_2018}, 
Zhang~\etal~\cite{Zhang_dehazing_2018}, 
Liu~\etal~\cite{GridDehazeNet_2019}, 3C~\cite{Ancuti_3C_TIP_2020} and 
Ancuti~\etal~\cite{Ancuti_NT_TIP_2020}.

Moreover, Fig.~\ref{fig:res_comp_crop1} shows the comparative detail insets of different scenes of the \textbf{NH-HAZE} dataset and the yielded results of the dehazing techniques previously mentioned.

On a close inspection we can observe that the well-know DCP~\cite{Dehaze_He_CVPR_2009} recovers quite well the image structure, but also amplifies the color shifting artifacts, while removing the varying hazy layers of the scene.

However, the operator introduced recently in~\cite{Ancuti_3C_TIP_2020} demonstrates that using 3C as a pre-processing step reduces significantly the color shifting introduced by the original DCP~\cite{Dehaze_He_CVPR_2009} and generates visually pleasing results for non-homogeneous hazy scenes.  

The results generated by the Berman~\etal~\cite{Berman_2016}, due to local airlight and transmission estimation strategy, present increased contrast, sharper edges and less color artifacts. The method of Ancuti~\etal~\cite{Ancuti_NT_TIP_2020} that also estimates locally the airlight, generates high contrast and vivid colors, but it tends to introduce a slight yellowish color-shifting for this set of images. Galdran~\etal~\cite{Galdran_2018} despite of the local strategy employed, presents some limitations to pleasantly restore the  local contrast.

The CNN-based techniques of Cai~\etal~\cite{Dehazenet_2016} and Liu~\etal~\cite{GridDehazeNet_2019} are limited to restore the contrast in the hazy regions mostly due to their strategy that assume homogeneous hazy scenes. On the other hand, the CNN-method of Zhang~\etal~\cite{Zhang_dehazing_2018} deals better with the variation of the haze in the scene. 

We draw the conclusion, that the CNN-based methods have a great potential and perform in general better than the other considered techniques~\cite{Dehaze_He_CVPR_2009,Berman_2016,Galdran_2018}. The main exception of the non-CNN techniques is the method of Ancuti~\etal~\cite{Ancuti_NT_TIP_2020}). These non-CNN techniques introduce higher color distortions compared with the CNN-based techniques and in general tend to introduce unnatural appearances of the results. In addition to the color shifting, these methods are prone to  amplify the structural artifacts and initial noise.

\textbf{NH-HAZE} has the main advantage to facilitate an objective quantitative evaluation based on the ground-truth haze-free images. This allows to identify the main limitations of the existing techniques while offering  important clues for future investigations.

In this work we perform an objective evaluation of the several image dehazing techniques based on \textbf{NH-HAZE}. Table~\ref{tabel_eval1} compares the output of different dehazing techniques with the ground-truth (haze-free) images based on PSNR and SSIM for the images shown in  Fig.~\ref{fig:res_comp1}. The structural similarity index (SSIM) compares local patterns of pixel intensities that have been normalized for luminance and contrast. The SSIM ranges in [-1,1], with maximum value 1 for two identical images.
In addition to Table~\ref{tabel_eval1}, Table~\ref{table_average} presents the average SSIM and PSNR values over the entire 55 scenes of the \textbf{NH-HAZE} dataset. From  these tables, we can conclude that in terms of structure and color restoration the methods of Zhang~\etal~\cite{Zhang_dehazing_2018} and Ancuti~\etal~\cite{Ancuti_NT_TIP_2020} perform the best on average when considering the SSIM and PSNR measures. As could be observed also visually, the other methods are less competitive both in terms of structure and color restoration.

Overall, none of the techniques performs better than others on all images. The low SSIM and PSNR values recorded for the analyzed techniques demonstrate once again that image dehazing is complex and that the non-homogeneity character of the haze poses additional challenges. 
\newline

\section*{Acknowledgments}
Part of this work has been supported by 2020 European Union Research and Innovation Horizon 2020 under the grant agreement Marie Sklodowska-Curie No 712949 (TECNIOspring PLUS), as well as the Agency for the Competitiveness of the Company of the Generalitat de Catalunya - ACCIO: TECSPR17-1-0054.

{\small
\bibliographystyle{ieee_fullname}
\bibliography{egbib}
}

\end{document}